# Advances in Boosting

**Robert E. Schapire**
AT&T Labs – Research
Shannon Laboratory
180 Park Avenue
Florham Park, NJ  07932

## Abstract

Boosting is a general method of generating many simple classification rules and combining them into a single, highly accurate rule. This paper reviews the AdaBoost boosting algorithm and some of its underlying theory, and then looks at some of the challenges of applying AdaBoost to bidding in complicated auctions and to human-computer spoken-dialogues systems.

## 1  INTRODUCTION

Boosting is a general method of generating many simple classification rules and combining them into a single, highly accurate rule. This paper reviews the boosting algorithm AdaBoost and some of its underlying theory, including extensions to boosting for logistic regression. A further overview of AdaBoost is given by Schapire [24].

Next, we look at two recent applications of boosting and some of the challenges that arose. The first application was as a tool for a trading agent that participated in a series of complicated, interacting auctions as part of the 2001 Trading Agent Competition. The agent required the ability to accurately predict prices in the auction, and moreover, to model their uncertainty. For this purpose, AdaBoost, which was originally designed only as a classification algorithm, was extended to handle such conditional density estimation problems.

The second application was to a human-computer spoken-dialogue system that is currently operational as a "help desk" for AT&T Labs' Natural Voices business. Here, the first challenge was a lack of adequate training data prior to deployment of the system. To compensate for the lack of data, AdaBoost was modified to incorporate prior knowledge and to balance such knowledge against available training data, however limited.

Another challenge came later in the development phase when so much data was flooding in that the annotators could not label it all. In this situation, we need to choose to label only the examples that we expect will be most informative. This can be done using AdaBoost by choosing to label the

Given: $(x_1, y_1), \ldots, (x_m, y_m)$ where $x_i \in X$, $y_i \in Y = \{-1, +1\}$
Initialize $D_1(i) = 1/m$.
For $t = 1, \ldots, T$:
- Train base learner using distribution $D_t$.
- Get base classifier $h_t : X \to \mathbb{R}$.
- Choose $\alpha_t \in \mathbb{R}$.
- Update:

$$D_{t+1}(i) = \frac{D_t(i) \exp(-\alpha_t y_i h_t(x_i))}{Z_t}$$

where $Z_t$ is a normalization factor (chosen so that $D_{t+1}$ will be a distribution).

Output the final classifier:

$$H(x) = \text{sign}\left(\sum_{t=1}^{T} \alpha_t h_t(x)\right).$$

Figure 1: The boosting algorithm AdaBoost.

examples whose predicitions have been assigned the lowest confidence.

## 2  ADABOOST

Working in Valiant's PAC (probably approximately correct) learning model [31], Kearns and Valiant [14, 15] were the first to pose the question of whether a "weak" learning algorithm that performs just slightly better than random guessing can be "boosted" into an arbitrarily accurate "strong" learning algorithm. Schapire [23] came up with the first provable polynomial-time boosting algorithm in 1989. A year later, Freund [9] developed a much more efficient boosting algorithm which, although optimal in a certain sense, nevertheless suffered like Schapire's algorithm from certain practical drawbacks.

The AdaBoost algorithm, introduced in 1995 by Freund and Schapire [10], solved many of the practical difficulties of the earlier boosting algorithms, and is the focus of this paper. Pseudocode for AdaBoost is given in Fig. 1 in the slightly generalized form given by Schapire and Singer [27]. The algorithm takes as input a training set $(x_1, y_1), \ldots, (x_m, y_m)$ where each $x_i$ belongs to some *domain* or *instance space* $X$, and each *label* $y_i$ is in some label set $Y$. For most of this paper, we assume $Y = \{-1, +1\}$; extensions to the multiclass case are given, for example, by Freund and Schapire [10], Schapire and Singer [27], and Allwein, Schapire and Singer [1].

AdaBoost calls a given *weak* or *base learning algorithm* repeatedly in a series of rounds $t = 1, \ldots, T$. One of the main ideas of the algorithm is to maintain a distribution or set of weights over the training set. The weight of this distribution on training example $i$ on round $t$ is denoted $D_t(i)$. Initially, all weights are set equally, but on each round, the weights of incorrectly classified examples are increased so that the base learner is forced to focus on the hard examples in the training set.

The base learner's job is to find a *base classifier* $h_t : X \to \mathbb{R}$ appropriate for the distribution $D_t$. In the simplest case, the range of each $h_t$ is binary, i.e., restricted to $\{-1, +1\}$; the base learner's job then is to minimize the *error*

$$\epsilon_t = \Pr_{i \sim D_t}\left[h_t(x_i) \neq y_i\right].$$

Once the base classifier $h_t$ has been received, AdaBoost chooses a parameter $\alpha_t \in \mathbb{R}$ that intuitively measures the importance that it assigns to $h_t$. In the figure, we have deliberately left the choice of $\alpha_t$ unspecified. For binary $h_t$, we typically set

$$\alpha_t = \tfrac{1}{2} \ln\left(\frac{1 - \epsilon_t}{\epsilon_t}\right) \qquad (1)$$

as in the original description of AdaBoost given by Freund and Schapire [10]. More on choosing $\alpha_t$ follows in Section 3. The distribution $D_t$ is then updated using the rule shown in the figure. The *final* or *combined classifier* $H$ is a weighted majority vote of the $T$ base classifiers where $\alpha_t$ is the weight assigned to $h_t$.

## 3 UNDERLYING THEORY

The most basic theoretical property of AdaBoost concerns its ability to reduce the training error, i.e., the fraction of mistakes on the training set. Specifically, Schapire and Singer [27], in generalizing a theorem of Freund and Schapire [10], show that the training error of the final classifier is bounded as follows:

$$\frac{1}{m}\left|\{i : H(x_i) \neq y_i\}\right| \leq \frac{1}{m}\sum_i \exp(-y_i f(x_i))$$
$$= \prod_t Z_t \qquad (2)$$

where henceforth we define

$$f(x) = \sum_t \alpha_t h_t(x) \qquad (3)$$

so that $H(x) = \text{sign}(f(x))$. (For simplicity of notation, we write $\sum_i$ and $\sum_t$ as shorthand for $\sum_{i=1}^m$ and $\sum_{t=1}^T$, respectively.) The inequality follows from the fact that $e^{-y_i f(x_i)} \geq 1$ if $y_i \neq H(x_i)$. The equality can be proved straightforwardly by unraveling the recursive definition of $D_t$.

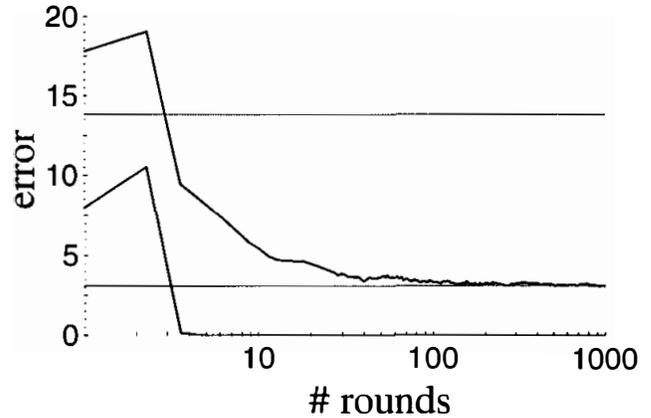

Figure 2: The training and test error curves (lower and upper curves, respectively) of the combined classifier as a function of the number of rounds of boosting, with C4.5 as the base learner, as reported by Schapire et al. [25] on the *letter* dataset. The horizontal lines indicate the test error rate of the base classifier as well as the test error of the final combined classifier.

Eq. (2) suggests that the training error can be reduced most rapidly (in a greedy way) by choosing $\alpha_t$ and $h_t$ on each round to minimize

$$Z_t = \sum_i D_t(i) \exp(-\alpha_t y_i h_t(x_i)). \qquad (4)$$

In the case of binary classifiers, this leads to the choice of $\alpha_t$ given in Eq. (1) and gives a bound on the training error of

$$\prod_t Z_t = \prod_t \left[2\sqrt{\epsilon_t(1 - \epsilon_t)}\right]$$
$$= \prod_t \sqrt{1 - 4\gamma_t^2} \leq \exp\left(-2\sum_t \gamma_t^2\right) \qquad (5)$$

where we define $\gamma_t = 1/2 - \epsilon_t$. This bound was first proved by Freund and Schapire [10]. Thus, if each base classifier is slightly better than random so that $\gamma_t \geq \gamma$ for some $\gamma > 0$, then the training error drops exponentially fast in $T$ since the bound in Eq. (5) is at most $e^{-2T\gamma^2}$.

Fig. 2 shows a sample run of boosting with both training and test error curves. As is here the case, and as has been observed by various authors, AdaBoost often tends not to overfit. Schapire et al. [25] provide a theoretical explanation for this tendency by proving bounds on boosting's generalization error that do not depend on the number of rounds of boosting, but rather depend on the "margins" (a measure of confidence of the predicted classifications) of the training examples which AdaBoost provably tends to increase with further rounds of training.

Eq. (2) points to the fact that, at heart, AdaBoost is a procedure for finding a linear combination $f$ of base classifiers which attempts to minimize

$$\sum_i \exp(-y_i f(x_i)) = \sum_i \exp\left(-y_i \sum_t \alpha_t h_t(x_i)\right). \tag{6}$$

Essentially, on each round, AdaBoost chooses $h_t$ (by calling the base learner) and then sets $\alpha_t$ to add one more term to the accumulating weighted sum of base classifiers in such a way that the sum of exponentials above will be maximally reduced. In other words, AdaBoost is doing a kind of steepest descent search to minimize Eq. (6) where the search is constrained at each step to follow coordinate directions (where we identify coordinates with the weights assigned to base classifiers). This view of boosting and its generalization are examined in considerable detail by Duffy and Helmbold [7], Mason et al. [20, 21] and Friedman [13]. See also Section 4.

Schapire and Singer [27] discuss the choice of $\alpha_t$ and $h_t$ in the case that $h_t$ is real-valued (rather than binary). In this case, $h_t(x)$ can be interpreted as a "confidence-rated prediction" in which the sign of $h_t(x)$ is the predicted label, while the magnitude $|h_t(x)|$ gives a measure of confidence. Here, Schapire and Singer advocate choosing $\alpha_t$ and $h_t$ so as to minimize $Z_t$ (Eq. (4)) on each round. Fig. 3 shows the results of experiments demonstrating that on some datasets, confidence-rated predictions can yield very substantial speed-ups in performance.

## 4 BOOSTING AND LOGISTIC REGRESSION

Classification generally is the problem of predicting the label $y$ of an example $x$ with the intention of minimizing the probability of an incorrect prediction. However, it is often useful to estimate the *probability* of a particular label. Friedman, Hastie and Tibshirani [12] suggested a method for using the output of AdaBoost to make reasonable estimates of such probabilities. Specifically, they suggested using a logistic function, and estimating

$$\Pr{}_f [y = +1 \mid x] = \frac{e^{f(x)}}{e^{f(x)} + e^{-f(x)}} \tag{7}$$

where, as usual, $f(x)$ is the weighted average of base classifiers produced by AdaBoost (Eq. (3)). The rationale for this choice is the close connection between the log loss (negative log likelihood) of such a model, namely,

$$\sum_i \ln\left(1 + e^{-2y_i f(x_i)}\right) \tag{8}$$

and the function that, we have already noted, AdaBoost attempts to minimize:

$$\sum_i e^{-y_i f(x_i)}. \tag{9}$$

Specifically, it can be verified that Eq. (8) is upper bounded by Eq. (9). In addition, if we add the constant $1 - \ln 2$ to Eq. (8) (which does not affect its minimization), then it can be verified that the resulting function and the one in Eq. (9) have identical Taylor expansions around zero up to second order; thus, their behavior near zero is very similar. Finally, it can be shown that, for any distribution over pairs $(x, y)$, the expectations

$$\mathrm{E}\left[\ln\left(1 + e^{-2yf(x)}\right)\right]$$

and

$$\mathrm{E}\left[e^{-yf(x)}\right]$$

are minimized by the same (unconstrained) function $f$, namely,

$$f(x) = \tfrac{1}{2} \ln\left(\frac{\Pr[y = +1 \mid x]}{\Pr[y = -1 \mid x]}\right).$$

Thus, for all these reasons, minimizing Eq. (9), as is done by AdaBoost, can be viewed as a method of approximately minimizing the negative log likelihood given in Eq. (8). Therefore, we may expect Eq. (7) to give a reasonable probability estimate.

Of course, as Friedman, Hastie and Tibshirani point out, rather than minimizing the exponential loss in Eq. (6), we could attempt instead to directly minimize the logistic loss in Eq. (8). To this end, they propose their LogitBoost algorithm. A different, more direct modification of AdaBoost for logistic loss was proposed by Collins, Schapire and Singer [4]. Following up on work by Kivinen and Warmuth [16] and Lafferty [17], they derive this algorithm using a unification of logistic regression and boosting based on Bregman distances. This work further connects boosting to the maximum-entropy literature, particularly the iterative-scaling family of algorithms [5, 6]. They also give unified proofs of convergence to optimality for a family of new and old algorithms, including AdaBoost, for both the exponential loss used by AdaBoost and the logistic loss used for logistic regression. See also the later work of Lebanon and Lafferty [18] who showed that logistic regression and boosting are in fact solving the same constrained optimization problem, except that in boosting, certain normalization constraints have been dropped.

For logistic regression, we attempt to minimize the loss function

$$\sum_i \ln\left(1 + e^{-y_i f(x_i)}\right) \tag{10}$$

which is the same as in Eq. (8) except for an inconsequential change of constants in the exponent. The modification of AdaBoost proposed by Collins, Schapire and Singer to handle this loss function is particularly simple. In AdaBoost, unraveling the definition of $D_t$ given in Fig. 1 shows that $D_t(i)$ is proportional (i.e., equal up to normalization) to

$$\exp\left(-y_i f_{t-1}(x_i)\right)$$

where we define

$$f_t(x) = \sum_{t'=1}^t \alpha_{t'} h_{t'}(x).$$

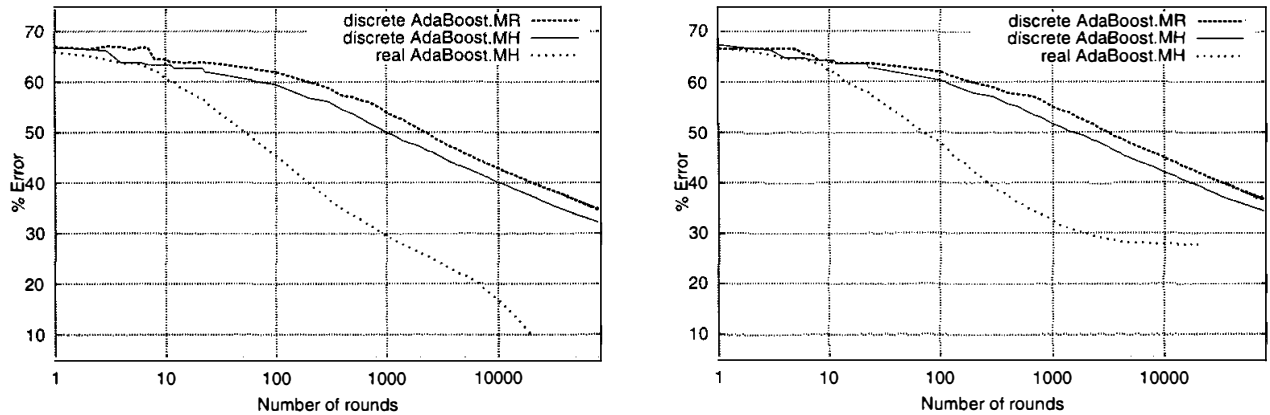

Figure 3: Comparison of the training (left) and test (right) error using three boosting methods on a six-class text classification problem from the TREC-AP collection, as reported by Schapire and Singer [27, 28]. Discrete AdaBoost.MH and discrete AdaBoost.MR are multiclass versions of AdaBoost that require binary ($\{-1,+1\}$-valued) base classifiers, while real AdaBoost.MH is a multiclass version that uses "confidence-rated" (i.e., real-valued) base classifiers.

To minimize the loss function in Eq. (10), the only necessary modification is to redefine $D_t(i)$ to be proportional to

$$\frac{1}{1+\exp\left(y_i f_{t-1}(x_i)\right)}.$$

A very similar algorithm is described by Duffy and Helmbold [7]. Note that in each case, the weight on the examples, viewed as a vector, is proportional to the negative gradient of the respective loss function. This is because both algorithms are doing a kind of functional gradient descent, an observation that is spelled out and exploited by Breiman [2], Duffy and Helmbold [7], Mason et al. [20, 21] and Friedman [13].

Having reviewed AdaBoost and some of its underlying theory, we turn now to two applications of boosting and how the theory helped us to face some of the challenges that arose.

## 5　MODELING AUCTION PRICE UNCERTAINTY

Boosting was recently applied to the design of agents participating in complicated auctions. When bidding for multiple interacting goods in simultaneous auctions, agents must be able to reason about uncertainty and make complex value assessments. For example, an agent bidding for a camera and flash may end up buying the flash and then not being able to find an affordable camera. Alternatively, if bidding for the same good in several auctions, it may purchase two flashes when only one was needed.

When bidding in any auction, it is important to be able to evaluate how much each item is worth to the agent. In interacting auctions, this also requires being able to predict the price of other items in the auction. For instance, in the example above, to determine the value of the camera, we need to guess the price of the flash; the amount that we are willing to spend on the camera then is the difference between the value of the camera-flash combination and the estimated price of the flash alone. Thus, a fundamental challenge when bidding for multiple goods is predicting the prices of all of the relevant goods before they are known.

To attack the price prediction problem, Stone et al.[30] proposed a machine-learning approach: gather examples of previous auctions and the prices paid in them, then use machine-learning methods to predict these prices. Moreover, for this strategy, we needed to be able to model the uncertainty associated with predicted prices; in other words, we needed to be able to sample from a predicted distribution of prices given the current state of the game. This can be viewed as a *conditional density estimation* problem, that is, a supervised learning problem in which the goal is to estimate the entire distribution of a real-valued label given a description of current conditions typically in the form of a feature vector.

The boosting-based conditional density estimation algorithm is described by Schapire et al.[29]. Briefly, in this setting, each example $x$ is now labeled with a real number $y$ whose distribution is to be estimated, conditional on $x$. To apply boosting, an algorithm designed for classification, we first discretize the range of $y$ by choosing a set of breakpoints $b_1, \ldots, b_k$. We then use the logistic regression version of boosting (see Section 4) to estimate, for each $j$, the probability that $y \geq b_j$ given $x$. Finally, we combine all of these estimated probabilities straightforwardly to obtain an estimate of $y$'s entire conditional distribution.

Stone et al. [30] successfully applied the algorithm to the problem of price prediction in auctions. It was implemented as part of ATTac-2001, a top-scoring agent[1] in the second Trading Agent Competition (TAC-01). In this competition, each agent had to bid simultaneously for multiple interacting goods. As observed above, the key challenge in such auctions is the modeling of uncertainty in the eventual prices of goods: with complete knowledge of eventual prices, there are direct methods for determining the optimal bids to place. The guiding principle in the design of ATTac-2001 was to have the agent model uncertainty and, to the greatest extent possible, analytically calculate optimal bids. ATTac-2001 used a predictive, data-driven approach to bidding based on expected marginal values of all available goods. The boosting-based price-predictor was at the heart of the algorithm.

---

[1]Top-scoring by one metric, and second place by another.

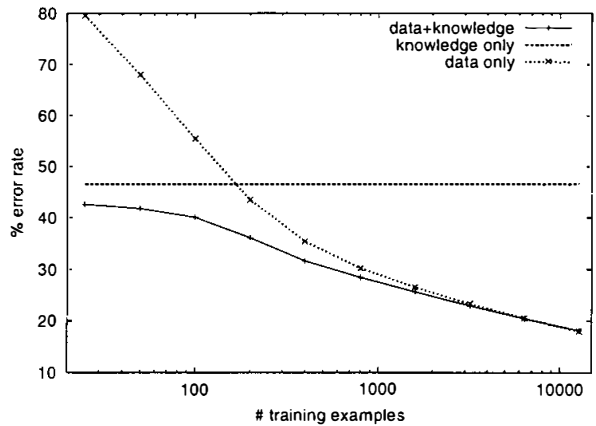

Figure 4: Comparison of test error rate using hand-crafted prior knowledge and data separately or together on the *AP-Titles* benchmark dataset, measured as a function of the number of training examples, as reported by Schapire et al. [26].

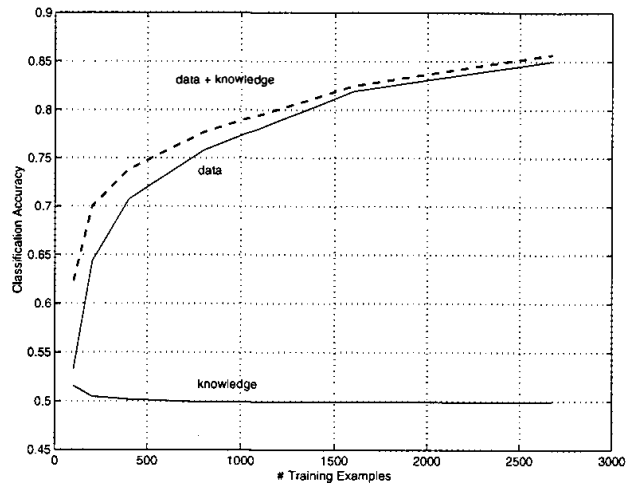

Figure 5: Comparison of performance using data and hand-crafted knowledge separately or together on the *HelpDesk* spoken-dialogue task, as reported by Rochery et al. [22].

## 6 SPOKEN-DIALOGUE SYSTEMS

Boosting has recently been used in the development of spoken-dialogue systems at AT&T. In these systems, a computer must formulate an appropriate response to the utterances of a telephone caller. The caller's utterance is converted to (noisy) text by an automatic speech recognizer. Next, the caller's meaning must be extracted to the extent that his or her utterance can be classified among a fixed set of categories; for instance, the utterance may be classified as a request for pricing information, a request to speak to a sales agent, etc. The computed classification is passed on to the dialogue manager which formulates a response that is converted to speech and played for the caller. Naturally, everything must happen in real time. The entire system is described in more detail by Di Fabbrizio et al. [8].

Boosting was applied specifically to the problem of constructing a classifier for extracting the meaning of caller utterances. For base classifiers, we used decision stumps, simple one-level decision trees that test for the presence of words or short phrases, as described by Schapire and Singer [28]. The classifier was trained from utterances that were labeled by human annotators. However, for rapid deployment, we were faced with a kind of "chicken and egg" problem: to train the classifier, we need data, but real data cannot be easily collected until the system is actually deployed. We solved this difficulty using the method described next which permitted us to use human-crafted knowledge to compensate for this initial dearth of data until enough could be collected following deployment.

## 7 INCORPORATING PRIOR KNOWLEDGE

Boosting, like many machine-learning methods, is entirely data-driven in the sense that the classifier it generates is derived exclusively from the evidence present in the training data itself. When data is abundant, this approach makes sense. However, in some applications, data may be severely limited, but there may be human knowledge that, in principle, might compensate for the lack of data.

In its standard form, boosting does not allow for the direct incorporation of such prior knowledge. Nevertheless, Schapire et al. [26] describe a modification of boosting that combines and balances human expertise with available training data. The aim of the approach is to allow the human's rough judgments to be refined, reinforced and adjusted by the statistics of the training data, but in a manner that does not permit the data to entirely overwhelm human judgments.

The first step in this approach is for a human expert to construct by hand a rule $p$ mapping each instance $x$ to an estimated probability $p(x) \in [0, 1]$ that is interpreted as the guessed probability that instance $x$ will appear with label $+1$. There are various methods for constructing such a function $p$, and the hope is that this difficult-to-build function need not be highly accurate for the approach to be effective.

Schapire et al.'s basic idea is to replace the logistic loss function in Eq. (10) with one that incorporates prior knowledge,

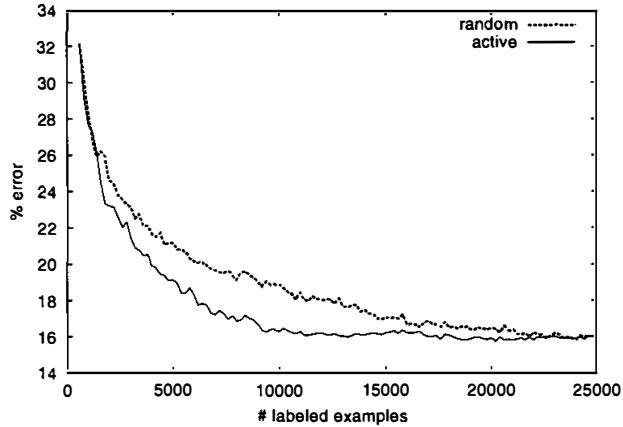

Figure 6: A preliminary experimental comparison of error rate obtained by AdaBoost, with decision stumps for base classifiers, on the *AP-Titles* dataset when examples are chosen for labeling either randomly or actively.

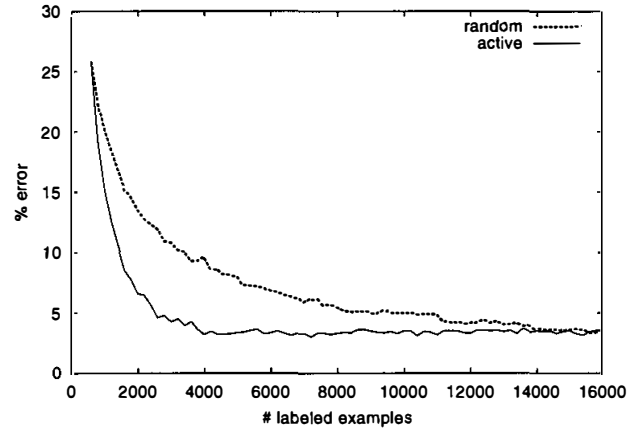

Figure 7: A preliminary experimental comparison of error rate obtained by AdaBoost, with C4.5 as the base learner, on the *letter* dataset when examples are chosen for labeling either randomly or actively.

namely,

$$\sum_i \ln\left(1 + e^{-y_i f(x_i)}\right) + \eta \sum_i \text{RE}\left(p(x_i) \,\|\, \frac{1}{1 + e^{-f(x_i)}}\right)$$

where $\text{RE}(p \,\|\, q) = p\ln(p/q) + (1-p)\ln((1-p)/(1-q))$ is binary relative entropy. The first term is the same as that in Eq. (10). The second term gives a measure of the distance from the model built by boosting to the human's model. Thus, we balance the conditional likelihood of the data against the distance from our model to the human's model. The relative importance of the two terms is controlled by the parameter $\eta > 0$.

Experiments using this method are shown in Figs. 4 and 5.

## 8　ACTIVE LEARNING

In the spoken-dialogue system described in Section 6, it is possible to cheaply gather a very large number of caller utterances, but annotation is quite expensive. Therefore, we want to actively choose to annotate only those unlabeled examples that are likely to be most informative.

This kind of active learning scenario was studied previously by various authors, including Lewis and Gale [19], Cohn, Atlas and Ladner [3], and Freund et al. [11]. Much of this work is based on the "uncertainty sampling" principle in which we ask to have labeled the examples that our classifier is most uncertain about. This same idea can be applied to boositng. Here, we can measure our confidence or certainty in a prediction by the magnitude of $f(x)$ (see Eq. (3)). So we can request labels for those examples for which $|f(x)|$ is smallest.

In other words, we assume we are given a large collection of unlabeled examples and a human annotator. We want to use the annotator's precious time as efficiently as possible. So we might first have the annotator label, say, 500 random examples to get things started and run AdaBoost on these. Next, we have the annotator label, say, the 200 unlabeled examples with lowest confidences $|f(x)|$. We repeat this process of running AdaBoost and labeling more low-confidence examples as many times as we can afford.

We can simulate this process on very large labeled datasets. Some preliminary results are given in Figs. 6 and 7 showing that, compared to choosing examples for labeling entirely at random, the reduction in the number of training examples that need to be labeled can be very substantial—up to a factor of four. Further theory is needed to understand why and when this approach is effective.

## 9　CONCLUSION

In this paper, we have outlined some of the theory underlying boosting, and how that theory has helped us to face the challenge of extending boosting to settings for which it was not originally designed. We have discussed boosting's application to the design of a trading agent for which we needed to estimate the conditional distribution of prices in a complicated auction. We also discussed boosting's application to a spoken-dialogue system where we needed to incorporate prior knowledge to compensate for initially insufficient data, and where we later needed to filter a large stream of unlabeled examples to select the ones whose labels are likely to be the most informative.

## References

[1] Erin L. Allwein, Robert E. Schapire, and Yoram Singer. Reducing multiclass to binary: A unifying approach for margin classifiers. *Journal of Machine Learning Research*, 1:113–141, 2000.

[2] Leo Breiman. Prediction games and arcing classifiers. *Neural Computation*, 11(7):1493–1517, 1999.

[3] David Cohn, Les Atlas, and Richard Ladner. Improving generalization with active learning. *Machine Learning*, 15(2):201–221, 1994.

[4] Michael Collins, Robert E. Schapire, and Yoram Singer. Logistic regression, AdaBoost and Bregman distances. *Machine Learning*, 48(1/2/3), 2002.

[5] J. N. Darroch and D. Ratcliff. Generalized iterative scaling for log-linear models. *The Annals of Mathematical Statistics*, 43(5):1470–1480, 1972.

[6] Stephen Della Pietra, Vincent Della Pietra, and John Lafferty. Inducing features of random fields. *IEEE Transactions Pattern Analysis and Machine Intelligence*, 19(4):1–13, April 1997.

[7] Nigel Duffy and David Helmbold. Potential boosters? In *Advances in Neural Information Processing Systems 11*, 1999.

[8] Giuseppe Di Fabbrizio, Dawn Dutton, Narendra Gupta, Barbara Hollister, Mazin Rahim, Giuseppe Riccardi, Robert Schapire, and Juergen Schroeter. AT&T help desk. In *7th International Conference on Spoken Language Processing*, 2002.

[9] Yoav Freund. Boosting a weak learning algorithm by majority. *Information and Computation*, 121(2):256–285, 1995.

[10] Yoav Freund and Robert E. Schapire. A decision-theoretic generalization of on-line learning and an application to boosting. *Journal of Computer and System Sciences*, 55(1):119–139, August 1997.

[11] Yoav Freund, H. Sebastian Seung, Eli Shamir, and Naftali Tishby. Selective sampling using the query by committee algorithm. *Machine Learning*, 28:133–168, 1997.

[12] Jerome Friedman, Trevor Hastie, and Robert Tibshirani. Additive logistic regression: A statistical view of boosting. *The Annals of Statistics*, 38(2):337–374, April 2000.

[13] Jerome H. Friedman. Greedy function approximation: A gradient boosting machine. *The Annals of Statistics*, 29(5), October 2001.

[14] Michael Kearns and Leslie G. Valiant. Learning Boolean formulae or finite automata is as hard as factoring. Technical Report TR-14-88, Harvard University Aiken Computation Laboratory, August 1988.

[15] Michael Kearns and Leslie G. Valiant. Cryptographic limitations on learning Boolean formulae and finite automata. *Journal of the Association for Computing Machinery*, 41(1):67–95, January 1994.

[16] Jyrki Kivinen and Manfred K. Warmuth. Boosting as entropy projection. In *Proceedings of the Twelfth Annual Conference on Computational Learning Theory*, pages 134–144, 1999.

[17] John Lafferty. Additive models, boosting and inference for generalized divergences. In *Proceedings of the Twelfth Annual Conference on Computational Learning Theory*, pages 125–133, 1999.

[18] Guy Lebanon and John Lafferty. Boosting and maximum likelihood for exponential models. In *Advances in Neural Information Processing Systems 14*, 2002.

[19] David Lewis and William Gale. Training text classifiers by uncertainty sampling. In *Seventeenth Annual International ACM SIGIR Conference on Research and Development in Information Retrieval*, 1994.

[20] Llew Mason, Jonathan Baxter, Peter Bartlett, and Marcus Frean. Functional gradient techniques for combining hypotheses. In *Advances in Large Margin Classifiers*. MIT Press, 1999.

[21] Llew Mason, Jonathan Baxter, Peter Bartlett, and Marcus Frean. Boosting algorithms as gradient descent. In *Advances in Neural Information Processing Systems 12*, 2000.

[22] M. Rochery, R. Schapire, M. Rahim, N. Gupta, G. Riccardi, S. Bangalore, H. Alshawi, and S. Douglas. Combining prior knowledge and boosting for call classification in spoken language dialogue. In *International Conference on Accoustics, Speech and Signal Processing*, 2002.

[23] Robert E. Schapire. The strength of weak learnability. *Machine Learning*, 5(2):197–227, 1990.

[24] Robert E. Schapire. The boosting approach to machine learning: An overview. In *MSRI Workshop on Nonlinear Estimation and Classification*, 2002.

[25] Robert E. Schapire, Yoav Freund, Peter Bartlett, and Wee Sun Lee. Boosting the margin: A new explanation for the effectiveness of voting methods. *The Annals of Statistics*, 26(5):1651–1686, October 1998.

[26] Robert E. Schapire, Marie Rochery, Mazin Rahim, and Narendra Gupta. Incorporating prior knowledge into boosting. In *Proceedings of the Nineteenth International Conference on Machine Learning*, 2002.

[27] Robert E. Schapire and Yoram Singer. Improved boosting algorithms using confidence-rated predictions. *Machine Learning*, 37(3):297–336, December 1999.

[28] Robert E. Schapire and Yoram Singer. BoosTexter: A boosting-based system for text categorization. *Machine Learning*, 39(2/3):135–168, May/June 2000.

[29] Robert E. Schapire, Peter Stone, David McAllester, Michael L. Littman, and János A. Csirik. Modeling auction price uncertainty using boosting-based conditional density estimation. In *Proceedings of the Nineteenth International Conference on Machine Learning*, 2002.

[30] Peter Stone, Robert E. Schapire, János A. Csirik, Michael L. Littman, and David McAllester. ATTac-2001: A learning, autonomous bidding agent. In *Workshop on Agent Mediated Electronic Commerce IV*, 2002.

[31] L. G. Valiant. A theory of the learnable. *Communications of the ACM*, 27(11):1134–1142, November 1984.